\documentclass[a4paper, 10pt, conference]{IEEEtran}
\usepackage[]{inputenc}
\usepackage{textcomp}

\makeatletter

\title{
\LARGE PHom-GeM: Persistent Homology for Generative Models
}

\IEEEoverridecommandlockouts
\usepackage{cite}
\usepackage{amsmath,amssymb,amsfonts}
\usepackage{mathtools}

\usepackage{epsfig}
\usepackage[mathscr]{euscript}
\usepackage{color}
\usepackage{booktabs} 
\usepackage{stmaryrd} 
\usepackage{bbm} 
\usepackage{blindtext, comment}
\usepackage{subcaption}
\usepackage{rotating} 
\usepackage{tikz}
\usepackage{standalone}
\usetikzlibrary{arrows, patterns}
\usetikzlibrary{decorations.pathreplacing,angles,quotes}
\usepackage{framed, standalone}
\usepackage{graphicx}
\usepackage{capt-of}
\usepackage{balance}
\usepackage{makecell} 
\usepackage[linesnumbered,lined,boxed,commentsnumbered,ruled,vlined]{algorithm2e}
\usepackage{setspace}
\usepackage{xcolor}
\def\BibTeX{{\rm B\kern-.05em{\sc i\kern-.025em b}\kern-.08em
    T\kern-.1667em\lower.7ex\hbox{E}\kern-.125emX}}

\begin{document}


\author{\IEEEauthorblockN{Jeremy Charlier}
\IEEEauthorblockA{\textit{University of Luxembourg} \\
Luxembourg, Luxembourg \\
jeremy.charlier@uni.lu}
\and
\IEEEauthorblockN{Radu State}
\IEEEauthorblockA{\textit{University of Luxembourg} \\
Luxembourg, Luxembourg \\
radu.state@uni.lu}
\and
\IEEEauthorblockN{Jean Hilger}
\IEEEauthorblockA{\textit{BCEE} \\
Luxembourg, Luxembourg \\
j.hilger@bcee.lu}
}

\maketitle

\begin{abstract}
Generative neural network models, including Generative Adversarial Network (GAN) and Auto-Encoders (AE), are among the most popular neural network models to generate adversarial data. 
The GAN model is composed of 
a generator that produces synthetic data and of a discriminator that discriminates between the generator's output and the true data.
AE consist 
of an encoder which maps the model distribution to a latent manifold and of a decoder which maps the latent manifold to a reconstructed distribution.
However, generative models are known to provoke chaotically scattered reconstructed distribution during their training, and consequently, incomplete generated adversarial distributions. 
Current distance measures fail to address this problem because they are not able to acknowledge the shape of the 
data manifold, i.e. its topological features, and the scale at which the manifold should be analyzed. 
We propose Persistent Homology for Generative Models, PHom-GeM, a new methodology to assess and measure the 
distribution of a generative model. 
PHom-GeM minimizes an objective 
function between the true and the reconstructed distributions and uses persistent homology, the study of the topological features of a space at different spatial resolutions, to compare the nature of the true and the generated distributions. 
Our experiments underline the potential of persistent homology for Wasserstein GAN in comparison to Wasserstein AE and Variational AE.
The experiments are conducted on a real-world data set particularly challenging for traditional distance measures and generative neural network models. 
PHom-GeM is the first methodology to propose a topological distance measure, the bottleneck distance, for generative models used to compare adversarial samples 
in the context of credit card transactions.
\end{abstract}


\section{Motivation} \label{sec::intro} 
The field of unsupervised learning has evolved significantly for the past few years thanks to adversarial networks publications. In \cite{goodfellow2014generative}, Goodfelow et al. introduced a Generative Adversarial Network framework called GAN. It is a class of generative models that play a competitive game between two networks in which the generator network must compete against an adversary according to a game theoretic scenario \cite{goodfellow2016deep}. The generator network produces samples from a noise distribution and its adversary, the discriminator network, tries to distinguish real samples from generated samples, respectively samples inherited from the training data and samples produced by the generator. Meanwhile, Variational Auto-Encoders (VAE) presented by Kingma et al. in \cite{kingma2013auto} have emerged as a well-established approach for synthetic data generation. Nevertheless, they might generate poor target distribution because of the KL divergence \cite{goodfellow2016deep}. We recall an AE is a neural network trained to copy its input manifold to its output manifold through a hidden layer. The encoder function sends the input space to the hidden space and the decoder function brings back the hidden space to the input space. By applying some of the Optimal Transport (OT) concepts gathered in \cite{villani2003topics} and noticeably, the Wasserstein distance, Arjovsky et al. introduced the Wasserstein GAN (WGAN) in \cite{arjovsky2017wasserstein}. It tries to avoid the mode collapse, a typical training convergence issue occurring between the generator and the discriminator. Gulrajani et al. further optimized the concept in \cite{gulrajani2017improved} by proposing a Gradient Penalty to Wasserstein GAN (GP-WGAN) capable to generate adversarial samples of higher quality. Similarly, Tolstikhin et al. in \cite{tolstikhin2017wasserstein} applied the same OT concepts to AE and, therefore, introduced Wasserstein AE (WAE), a new type of AE generative model, that avoids the use of the KL divergence.  \\ 

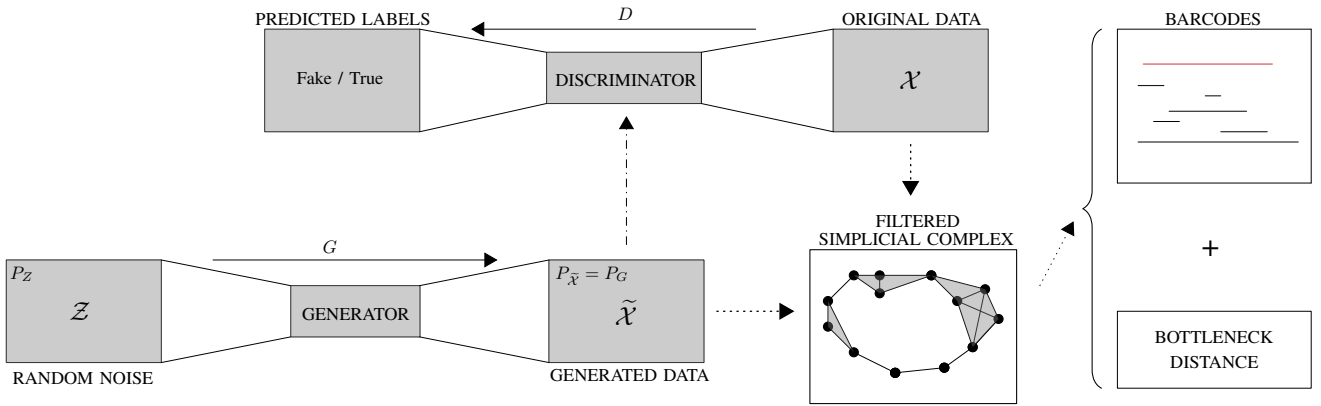
\begin{figure*}[t] 
  \begin{center} 
  \definecolor{dtsfsf}{rgb}{0.8274509803921568,0.1843137254901961,0.1843137254901961}
\definecolor{rvwvcq}{rgb}{0.08235294117647059,0.396078431372549,0.7529411764705882}

\begin{tikzpicture}[scale=0.68, transform shape, line cap=round,line join=round,>=triangle 45,x=1cm,y=1cm]

\draw [line width=0.25pt] (8.5,7.5) rectangle (11.5,9.5);
\fill [gray, opacity=.4] (8.5,7.5) rectangle (11.5,9.5);
\draw (9.65,8.8) node[anchor=north west] {\Large{$\mathcal{X}$}};

\draw [line width=0.25pt] (2.9549,8.05) rectangle (5.95,9.05);
\fill [gray, opacity=.4] (2.9549,8.05) rectangle (5.95,9.05);

\draw [line width=0.25pt] (-2.5,7.5) rectangle (0.5,9.5);
\fill [gray, opacity=.4] (-2.5,7.5) rectangle (0.5,9.5);

\draw [line width=0.25pt] (0.5,9.5)-- (2.95,9.05);
\draw [line width=0.25pt] (0.5,7.5)-- (2.95,8.05);

\draw [line width=0.25pt] (5.95,9.05)-- (8.5,9.5);
\draw [line width=0.25pt] (5.95,8.05)-- (8.5,7.5);

\draw [line width=0.25pt] (-7.5,3) rectangle (-4.5,5);
\fill [gray, opacity=.4] (-7.5,3) rectangle (-4.5,5);

\draw [line width=0.25pt] (-1.9951,3.5) rectangle (0.5049,4.5);
\fill [gray, opacity=.4] (-1.9951,3.5) rectangle (0.5049,4.5);
\draw (-6.4,4.3) node[anchor=north west] {\Large{$\mathcal{Z}$}}; 

\draw [line width=0.25pt] (3,3) rectangle (6,5);
\fill [gray, opacity=.4] (3,3) rectangle (6,5);

\draw [line width=0.25pt] (-4.5,5)-- (-2,4.5);
\draw [line width=0.25pt] (-4.5,3)-- (-2,3.5);

\draw [line width=0.25pt] (0.5,4.5)-- (3,5);
\draw [line width=0.25pt] (0.5,3.5)-- (3,3);

\draw [line width=0.25pt] (14,9.5)-- (14,6.5);
\draw [line width=0.25pt] (17.75,6.5)-- (14,6.5);
\draw [line width=0.25pt] (17.75,6.5)-- (17.75,9.5);
\draw [line width=0.25pt] (17.75,9.5)-- (14,9.5);
\draw [line width=0.25pt,color=dtsfsf] (14.5,8.82)-- (17,8.82);
\draw [line width=0.25pt] (14.4,8.4)-- (14.9,8.4);
\draw [line width=0.25pt] (15.7,8.2)-- (16,8.2);
\draw [line width=0.25pt] (15,7.9)-- (16.5,7.9);
\draw [line width=0.25pt] (14.7,7.7)-- (15.2,7.7);
\draw [line width=0.25pt] (16,7.5)-- (16.9,7.5);
\draw [line width=0.25pt] (14.4,7.3)-- (17.5,7.3);

%
%
%
%
%
%
%

\draw [line width=0.25pt] (9.4,4.35)-- (8.9,4.7);
\draw [line width=0.25pt] (8.4,4.2)-- (8.9,3.2);
\draw [line width=0.25pt] (8.9,4.7)-- (8.4,4.2);
\draw [line width=0.25pt] (8.9,3.2)-- (9.7,2.8);
\draw [line width=0.25pt] (9.7,2.8)-- (9.7,2.8);
\draw [line width=0.25pt] (9.7,2.8)-- (9.7,2.8);
\draw [line width=0.25pt] (9.7,2.8)-- (9.7,2.8);
\draw [line width=0.25pt] (9.7,2.8)-- (10.65,2.9);
\draw [line width=0.25pt] (10.65,2.9)-- (10.65,2.9);

\draw [line width=0.25pt] (11.2,3.3)-- (11.2,3.3);
\draw [line width=0.25pt] (11.2,3.3)-- (10.9,4.2);
\draw [line width=0.25pt] (10.9,4.2)-- (10.4,4.7);
\draw [line width=0.25pt] (10.4,4.7)-- (10.4,4.7);
\draw [line width=0.25pt] (10.4,4.7)-- (9.4,4.35);
\draw [line width=0.25pt] (8.4,4.2)-- (8.4,3.7);
\draw [line width=0.25pt] (8.9,3.2)-- (8.4,3.7);

\draw [line width=0.25pt] (10.65,2.9)-- (10.65,2.9);
\draw [line width=0.25pt] (10.65,2.9)-- (11.2,3.3);

\draw [line width=0.25pt] (11.44,4.43)-- (11.7,3.85);
\draw [line width=0.25pt] (11.7,3.85)-- (11.2,3.3);
\draw [line width=0.25pt] (11.7,3.85)-- (11.2,3.3);
\draw [line width=0.25pt] (11.2,3.3)-- (11.44,4.43);
\draw [line width=0.25pt] (11.44,4.43)-- (10.4,4.7);
\draw [line width=0.25pt] (10.9,4.2)-- (11.44,4.43);
\draw [line width=0.25pt] (10.9,4.2)-- (11.7,3.85);
\draw [line width=0.25pt] (9.4,4.35)-- (9.4,4.7);
\draw [line width=0.25pt] (9.4,4.7)-- (10.4,4.7);
\draw [line width=0.25pt] (9.4,4.7)-- (8.9,4.7);

\draw [fill=black] (9.4,4.35) circle (2.5pt);
\draw [fill=black] (8.9,4.7) circle (2.5pt);
\draw [fill=black] (8.4,4.2) circle (2.5pt);
\draw [fill=black] (8.9,3.2) circle (2.5pt);
\draw [fill=black] (9.7,2.8) circle (2.5pt);
\draw [fill=black] (9.7,2.8) circle (2.5pt);
\draw [fill=black] (9.7,2.8) circle (2.5pt);
\draw [fill=black] (9.7,2.8) circle (2.5pt);
\draw [fill=black] (10.65,2.9) circle (2.5pt);
\draw [fill=black] (10.65,2.9) circle (2.5pt);

\draw [fill=black] (11.2,3.3) circle (2.5pt);
\draw [fill=black] (11.2,3.3) circle (2.5pt);
\draw [fill=black] (10.9,4.2) circle (2.5pt);
\draw [fill=black] (10.4,4.7) circle (2.5pt);
\draw [fill=black] (10.4,4.7) circle (2.5pt);
\draw [fill=black] (8.4,3.7) circle (2.5pt);

\draw [fill=black] (10.65,2.9) circle (2.5pt);
\draw [fill=black] (11.44,4.43) circle (2.5pt);
\draw [fill=black] (11.7,3.85) circle (2.5pt);
\draw [fill=black] (9.4,4.7) circle (2.5pt);

\fill[line width=0.25pt,color=gray,fill=gray,fill opacity=0.4] (9.4,4.35) -- (8.9,4.7) -- (9.4,4.7) -- cycle;
\fill[line width=0.25pt,color=gray,fill=gray,fill opacity=0.4] (9.4,4.35) -- (10.4,4.7) -- (9.4,4.7) -- cycle;
\fill[line width=0.25pt,color=gray,fill=gray,fill opacity=0.4] (8.4,4.2) -- (8.9,3.2) -- (8.4,3.7) -- cycle;
\fill[line width=0.25pt,color=gray,fill=gray,fill opacity=0.4] (10.4,4.7) -- (11.44,4.43)  -- (11.7,3.85) -- (11.2,3.3) -- (11.2,3.3) -- (10.9,4.2) -- cycle;

\draw (8.05,5.2) node (v1) {} -- (8.05,2.2) -- (12.05,2.2) -- (12.05,5.2) -- cycle;

\draw (-2.8,9.95) node[anchor=north west] {{PREDICTED LABELS}};
\draw (-2.0,8.8) node[anchor=north west] {Fake / True };
\draw (3.0,8.75) node[anchor=north west] {{DISCRIMINATOR}};
\draw (8.55,9.95) node[anchor=north west] {{ORIGINAL DATA}};

\draw (-7.50,2.95) node[anchor=north west] {{RANDOM NOISE}};
\draw (-1.90,4.2) node[anchor=north west] {{GENERATOR}};
\draw (2.9,3) node[anchor=north west] {{GENERATED DATA}};

\draw (14.8,9.95) node[anchor=north west] {BARCODES};
\draw (9.20,6) node[anchor=north west] {FILTERED};
\draw (8.1,5.65) node[anchor=north west] {SIMPLICIAL COMPLEX};

\draw (-7.55,5) node[anchor=north west] {$P_Z$};
\draw (4.2,10.05) node[anchor=north west] {$D$}; 
\draw (-1.5,5.5) node[anchor=north west] {$G$}; 
\draw (3,5) node[anchor=north west] {$P_\mathcal{\widetilde{X}}=P_G$};
\draw (4.15,4.35) node[anchor=north west] {\Large{$\mathcal{\widetilde{X}}$}};

\draw [line width=0.25pt, dashdotted, -triangle 60] (4.5,5.3) -- (4.5,7.8);
\draw [line width=0.25pt, -triangle 60] (-3.5,5) -- (2,5);
\draw [line width=0.25pt, -triangle 60] (7,9.5) -- (1.5,9.5);

\draw [line width=0.5pt, dotted, -triangle 60] (10,7.25) -- (10,6.25);
\draw [line width=0.5pt, dotted, -triangle 60] (6.25,4) -- (7.75,4);

\draw[decoration={brace, raise=5pt, amplitude=7pt},decorate] (13.95,2.5) -- (13.95,9.5);

\draw [line width=0.25pt, -triangle 60, dotted] (12.5,4.5)-- (13.15,5.75);

\draw [line width=0.25pt] (14,2.5) rectangle (17.75,4);
\draw (14.6,3.75) node[anchor=north west] {BOTTLENECK};
\draw (14.9,3.25) node[anchor=north west] {DISTANCE};
\draw (15.8,5.5) node[anchor=north] {\LARGE{+}};

\end{tikzpicture} 
    \caption{In PHom-GeM applied to GAN, the generative model $G$ generates fake samples $\widetilde{X} \in \mathcal{\widetilde{X}}$ based on the samples $Z\in\mathcal{Z}$ from the prior random distribution $P_Z$. Then, the discriminator model $D$ tries to differentiate the fake samples $\widetilde{X}$ from the true samples $X\in\mathcal{X}$. The original manifold $\mathcal{X}$ and the generated manifold $\mathcal{\widetilde{X}}$ are transformed independently into metric space sets to obtain filtered simplicial complex. It leads to the description of topological features, summarized by the barcodes, to compare the respective topological representation of the true data distribution $P_X$ and the generative model distribution $P_G$. } \label{fig::phomgen} 
  \end{center} 
\end{figure*}

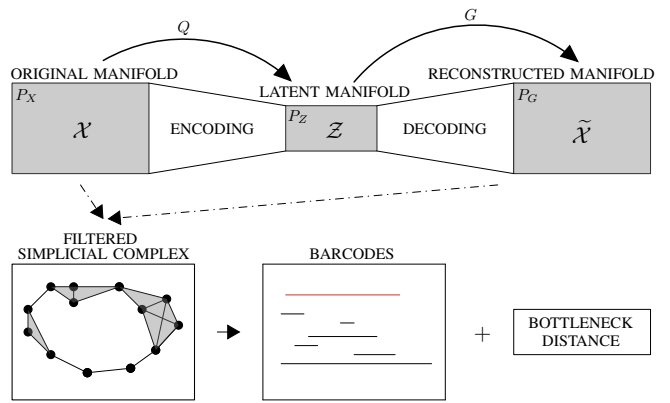
\begin{figure}[t]
	\begin{center}
	\definecolor{dtsfsf}{rgb}{0.8274509803921568,0.1843137254901961,0.1843137254901961}
\definecolor{rvwvcq}{rgb}{0.08235294117647059,0.396078431372549,0.7529411764705882}

\begin{tikzpicture}[scale=0.60, transform shape, line cap=round,line join=round,>=triangle 45,x=1cm,y=1cm]

\draw [line width=0.25pt] (-7,1) rectangle (-4,3);
\fill [gray, opacity=.4] (-7,1) rectangle (-4,3);
\draw (-5.8,2.25) node[anchor=north west] {\Large{$\mathcal{X}$}};

\draw [line width=0.25pt] (-1,1.5) rectangle (1,2.5);
\fill [gray, opacity=.4] (-1,1.5) rectangle (1,2.5);
\draw (-0.25,2.25) node[anchor=north west] {\Large{$\mathcal{Z}$}}; 

\draw [line width=0.25pt] (4,1) rectangle (7,3);
\fill [gray, opacity=.4] (4,1) rectangle (7,3);

\draw [line width=0.25pt] (-4,3)-- (-1,2.5);
\draw [line width=0.25pt] (-4,1)-- (-1,1.5);

\draw [line width=0.25pt] (1,2.5)-- (4,3);
\draw [line width=0.25pt] (1,1.5)-- (4,1);

\draw  [draw=black, line width=0.2mm, ->]  (0.5,3.025) arc (140:50:3.5);
\draw (2.8,4.75) node[anchor=north west] {$G$};

\draw  [draw=black, line width=0.2mm, ->]  (-5,3.5) arc (120.00:47:3.5);
\draw (-3.5,4.5) node[anchor=north west] {$Q$};

\draw [line width=0.25pt] (-1.5,-1)-- (-1.5,-4);
\draw [line width=0.25pt] (2.5,-4)-- (-1.5,-4);
\draw [line width=0.25pt] (2.5,-4)-- (2.5,-1);
\draw [line width=0.25pt] (2.5,-1)-- (-1.5,-1);
\draw [line width=0.25pt,color=dtsfsf] (-1,-1.68)-- (1.5,-1.68);
\draw [line width=0.25pt] (-1.1,-2.1)-- (-0.6,-2.1);
\draw [line width=0.25pt] (0.2,-2.3)-- (0.5,-2.3);
\draw [line width=0.25pt] (-0.5,-2.6)-- (1,-2.6);
\draw [line width=0.25pt] (-0.8,-2.8)-- (-0.3,-2.8);
\draw [line width=0.25pt] (0.5,-3)-- (1.4,-3);
\draw [line width=0.25pt] (-1.1,-3.2)-- (2.2,-3.2);

\draw [line width=0.25pt, -triangle 60] (-2.5,-2.5)-- (-2,-2.5);
\draw (3.0,-2.25) node[anchor=north west] {\Large\textbf{$+$}};

\draw [line width=0.25pt, dashdotted, -triangle 60] (-5.5,0.75)-- (-5,0);
\draw [line width=0.25pt, dashdotted, -triangle 60] (3.65,0.75)-- (-4.9,0);

\draw [line width=0.25pt] (-5.65,-1.85)-- (-6.15,-1.5);
\draw [line width=0.25pt] (-6.65,-2)-- (-6.15,-3);
\draw [line width=0.25pt] (-6.15,-1.5)-- (-6.65,-2);
\draw [line width=0.25pt] (-6.15,-3)-- (-5.35,-3.4);
\draw [line width=0.25pt] (-5.35,-3.4)-- (-5.35,-3.4);
\draw [line width=0.25pt] (-5.35,-3.4)-- (-5.35,-3.4);
\draw [line width=0.25pt] (-5.35,-3.4)-- (-5.35,-3.4);
\draw [line width=0.25pt] (-5.35,-3.4)-- (-4.4,-3.3);
\draw [line width=0.25pt] (-4.4,-3.3)-- (-4.4,-3.3);

\draw [line width=0.25pt] (-3.85,-2.9)-- (-3.85,-2.9);
\draw [line width=0.25pt] (-3.85,-2.9)-- (-4.15,-2);
\draw [line width=0.25pt] (-4.15,-2)-- (-4.65,-1.5);
\draw [line width=0.25pt] (-4.65,-1.5)-- (-4.65,-1.5);
\draw [line width=0.25pt] (-4.65,-1.5)-- (-5.65,-1.85);
\draw [line width=0.25pt] (-6.65,-2)-- (-6.65,-2.5);
\draw [line width=0.25pt] (-6.15,-3)-- (-6.65,-2.5);

\draw [line width=0.25pt] (-4.4,-3.3)-- (-4.4,-3.3);
\draw [line width=0.25pt] (-4.4,-3.3)-- (-3.85,-2.9);

\draw [line width=0.25pt] (-3.61,-1.77)-- (-3.35,-2.35);
\draw [line width=0.25pt] (-3.35,-2.35)-- (-3.85,-2.9);
\draw [line width=0.25pt] (-3.35,-2.35)-- (-3.85,-2.9);
\draw [line width=0.25pt] (-3.85,-2.9)-- (-3.61,-1.77);
\draw [line width=0.25pt] (-3.61,-1.77)-- (-4.65,-1.5);
\draw [line width=0.25pt] (-4.15,-2)-- (-3.61,-1.77);
\draw [line width=0.25pt] (-4.15,-2)-- (-3.35,-2.35);
\draw [line width=0.25pt] (-5.65,-1.85)-- (-5.65,-1.5);
\draw [line width=0.25pt] (-5.65,-1.5)-- (-4.65,-1.5);
\draw [line width=0.25pt] (-5.65,-1.5)-- (-6.15,-1.5);

\draw [fill=black] (-5.65,-1.85) circle (2.5pt);
\draw [fill=black] (-6.15,-1.5) circle (2.5pt);
\draw [fill=black] (-6.65,-2) circle (2.5pt);
\draw [fill=black] (-6.15,-3) circle (2.5pt);
\draw [fill=black] (-5.35,-3.4) circle (2.5pt);
\draw [fill=black] (-5.35,-3.4) circle (2.5pt);
\draw [fill=black] (-5.35,-3.4) circle (2.5pt);
\draw [fill=black] (-5.35,-3.4) circle (2.5pt);
\draw [fill=black] (-4.4,-3.3) circle (2.5pt);
\draw [fill=black] (-4.4,-3.3) circle (2.5pt);

\draw [fill=black] (-3.85,-2.9) circle (2.5pt);
\draw [fill=black] (-3.85,-2.9) circle (2.5pt);
\draw [fill=black] (-4.15,-2) circle (2.5pt);
\draw [fill=black] (-4.65,-1.5) circle (2.5pt);
\draw [fill=black] (-4.65,-1.5) circle (2.5pt);
\draw [fill=black] (-6.65,-2.5) circle (2.5pt);

\draw [fill=black] (-4.4,-3.3) circle (2.5pt);
\draw [fill=black] (-3.61,-1.77) circle (2.5pt);
\draw [fill=black] (-3.35,-2.35) circle (2.5pt);
\draw [fill=black] (-5.65,-1.5) circle (2.5pt);

\fill[line width=0.25pt,color=gray,fill=gray,fill opacity=0.4] (-5.65,-1.85) -- (-6.15,-1.5) -- (-5.65,-1.5) -- cycle;
\fill[line width=0.25pt,color=gray,fill=gray,fill opacity=0.4] (-5.65,-1.85) -- (-4.65,-1.5) -- (-5.65,-1.5) -- cycle;
\fill[line width=0.25pt,color=gray,fill=gray,fill opacity=0.4] (-6.65,-2) -- (-6.15,-3) -- (-6.65,-2.5) -- cycle;
\fill[line width=0.25pt,color=gray,fill=gray,fill opacity=0.4] (-4.65,-1.5) -- (-3.61,-1.77)  -- (-3.35,-2.35) -- (-3.85,-2.9) -- (-3.85,-2.9) -- (-4.15,-2) -- cycle;

\draw (-7,-1) node (v1) {} -- (-7,-4) -- (-3,-4) -- (-3,-1) -- cycle;

\draw (-0.6,-0.525) node[anchor=north west] {BARCODES};
\draw (-6.00,-0.2) node[anchor=north west] {FILTERED};
\draw (-7.00,-0.525) node[anchor=north west] {SIMPLICIAL COMPLEX};
\draw (-3.65,2.25) node[anchor=north west] {ENCODING};
\draw (1.45,2.25) node[anchor=north west] {DECODING};
\draw (-7.15,3.45) node[anchor=north west] {ORIGINAL MANIFOLD};
\draw (2.0,3.45) node[anchor=north west] {RECONSTRUCTED MANIFOLD};
\draw (-1.7,3.05) node[anchor=north west] {LATENT MANIFOLD};
\draw (4.2,-2.05) node[anchor=north west] {BOTTLENECK};
\draw (4.5,-2.45) node[anchor=north west] {DISTANCE};
\draw [line width=0.25pt] (4,-2) rectangle (7,-3);

\draw (-7.05,3) node[anchor=north west] {$P_X$};
\draw (-1.075,2.525) node[anchor=north west] {$P_Z$}; 
\draw (3.95,3) node[anchor=north west] {$P_G$};
\draw (5.15,2.3) node[anchor=north west] {\Large{$\mathcal{\widetilde{X}}$}};

\end{tikzpicture}
    \caption{In PHom-GeM applied to AE, the generative model $G$, the decoder, is used to generate fake samples $\widetilde{X} \in \mathcal{\widetilde{X}}$ based on the samples $Z\in\mathcal{Z}$ from a prior random distribution $P_Z$. Afterward, the original manifold $\mathcal{X}$ and the generated manifold $\mathcal{\widetilde{X}}$ are both transformed independently into metric space sets to obtain filtered simplicial complex. As for PHom-GeM applied to GAN, it leads to the description of topological features, summarized by the barcodes, to compare the respective topological representation of the true data distribution $P_X$ and the generative model distribution $P_G$.}
    \label{fig::autoencoder}
	\end{center}
\end{figure}

Nonetheless, the description of the distribution $P_G$ of the generative models, which involves the description of the generated scattered data points \cite{bengio2013generalized} based on the distribution $P_X$ of the original manifold $\mathcal{X}$, is very difficult using traditional distance measures, such as the Fr\'echet Inception Distance \cite{tolstikhin2017wasserstein}. We highlight the distribution and the manifold notations in figure  \ref{fig::phomgen} for GAN and in figure \ref{fig::autoencoder} for AE. Effectively, traditional distance measures are not able to acknowledge the shapes of the data manifolds and the scale at which the manifold should be analyzed. However, persistent homology \cite{Edelsbrunner2002tps,zomorodian2005computing} is specifically designed to highlight the topological features of the data \cite{chazal2017tda}. Therefore, building upon persistent homology, Wasserstein distance \cite{bousquet2017optimal} and generative models \cite{tolstikhin2017wasserstein}, our main contribution is to propose qualitative and quantitative ways to evaluate the scattered generated distributions and the performance of the generative models. \\

In this work we describe the persistent homology features of the generated model $G$ while minimizing the OT function $W_c (P_X, P_G)$ for a squared cost $c(x,y)=||x-y ||_2^2$ where $P_X$ is the model distribution of the data contained in the manifold $\mathcal{X}$, and $P_G$ the distribution of the generative model capable of generating adversarial samples. Our contributions are summarized below: \\

\begin{itemize} 
  \item A persistent homology procedure for generative models, including GP-WGAN, WGAN, WAE and VAE, which we call PHom-GeM to highlight the topological properties of the generated distributions of the data for different spatial resolutions. The objective is a persistent homology description of the generated data distribution $P_G$ following the generative model $G$. 
  \item A distance measure for persistence diagrams, the bottleneck distance, applied to generative models to compare quantitatively the true and the target distributions on any data set. We measure the shortest distance for which there exists a perfect matching between the points of the two persistence diagrams. A persistence diagram is a stable summary representation of topological features of simplicial complex, a collection of vertices, associated to the data set.
  \item Finally, we propose the first application of algebraic topology and generative models on a public data set containing credit card transactions, particularly challenging for this type of models and traditional distance measures. 
\end{itemize}

The paper is structured as follows. In section \ref{sec::propmethod}, we review the optimized GP-WGAN and WAE formulations using OT derived by Gulrajani et al. in \cite{gulrajani2017improved} and Tolstikhin et al. in \cite{tolstikhin2017wasserstein}, respectively. By using persistence homology, we are able to compare the topological properties of the original distribution $P_X$ and the generated distribution $P_G$. 
We highlight experimental results in section \ref{sec::exp} and we conclude in section \ref{sec::ccl} by addressing promising directions for future work.

\section{Proposed Method} \label{sec::propmethod} 
Our method computes the persistence homology of both the true manifold $X\in\mathcal{X}$ and the generated manifold $\widetilde{X} \in \mathcal{\widetilde{X}}$ following the generative model $G$ based on the minimization of the optimal transport cost $W_c(P_X, P_G)$. In the resulting topological problem, the points of the manifolds are transformed to a metric space set for which a Vietoris-Rips simplicial complex filtration is applied (see definition 2). PHom-GeM achieves simultaneously two main goals: it computes the birth-death of the pairing generators of the iterated inclusions while measuring the bottleneck distance between persistence diagrams of the manifolds of the generative models.


\subsection{Optimal Transport and Dual Formulation} 
Following the description of the optimal transport problem \cite{villani2003topics} and relying on the Kantorovich-Rubinstein duality, the Wasserstein distance is computed as

\begin{equation} 
  W_c(P_X, P_G) = \sup_{f \in \mathcal{F}_L} \mathbb{E}_{X\sim P_X}[f(X)] - \mathbb{E}_{Y\sim P_G}[f(Y)] 
\end{equation} 

where $(\mathcal{X}, d)$ is a metric space, $\mathcal{P}(X\sim P_X, Y\sim P_G)$ is a set of all joint distributions $(X, Y)$ with marginals $P_X$ and $P_G$ respectively and $\mathcal{F}_L$ is the class of all bounded 1-Lipschitz functions on $(\mathcal{X}, d)$. \\

\subsection{Gradient Penalty Wasserstein GAN (GP-WGAN)} 
As described in \cite{gulrajani2017improved}, the GP-WGAN objective loss function with gradient penalty is expressed such that  

\begin{equation} \label{eq::DWGAN} 
\begin{aligned} 
L = &\underset{\widetilde{X}\sim P_G}{\mathbb{E}}[f(\widetilde{X})] - \underset{X\sim P_X}{\mathbb{E}}[f(X)] \\ 
&+ \lambda \underset{\widehat{X}\sim P_{\widehat{X}}}{\mathbb{E}} [(||\nabla_{\widehat{X}} f(\widehat{X})||_2 - 1)^2] 
\end{aligned} 
\end{equation} 

where $f$ is the set of 1-Lipschitz functions on $(\mathcal{X}, d)$, $P_X$ the original data distribution, $P_G$ the generative model distribution implicitly defined by $\widetilde{X}=G(Z), Z\sim p(Z)$. The input $Z$ to the generator is sampled from a noise distribution such as a uniform distribution. $P_{\widehat{X}}$ defines the uniform sampling along straight lines between pairs of points sampled from the data distribution $P_X$ and the generative distribution $P_G$. A penalty on the gradient norm is enforced for random samples $\widehat{X}\sim P_{\widehat{X}}$. For further details, we refer to \cite{gulrajani2017improved} and \cite{arjovsky2017wasserstein}. \\

\subsection{Wasserstein Auto-Encoders} 
As described in \cite{tolstikhin2017wasserstein}, the WAE objective function is expressed such that  

\begin{equation} \label{eq::DWAE} 
  \begin{split} 
    D_{\text{WAE}}(P_X, P_G) := & \underset{Q(Z|X)\in\mathcal{Q}}{\inf} \mathbb{E}_{P_X} \mathbb{E}_{Q(Z|X)} [c(X, G(Z))] \\ 
    & \quad + \lambda \mathcal{D}_Z(Q_Z, P_Z) 
  \end{split} 
\end{equation} 

where $c(X, G(Z)): \mathcal{X}\times \mathcal{X}\rightarrow\mathcal{R}_+$ is any measurable cost function. In our experiments, we use a square cost function $c(x,y)=||x-y||_2^2$ for data points $x,y \in \mathcal{X}$. $G(Z)$ denotes the sending of $Z$ to $X$ for a given map $G:\mathcal{Z}\rightarrow \mathcal{X}$. $Q$, and $G$, are any nonparametric set of probabilistic encoders, and decoders respectively. \\ 

We use the Maximum Mean Discrepancy (MMD) for the penalty $\mathcal{D}_Z(Q_Z, P_Z)$ for a positive-definite reproducing kernel $k:\mathcal{Z}\times\mathcal{Z}\rightarrow\mathcal{R}$ 

\begin{equation} 
\begin{split} 
  \mathcal{D}_Z(P_Z, Q_Z) := & \text{MMD}_k(P_Z, Q_Z) \\
  = & || \int_\mathcal{Z}k(z, .) dP_Z(z) - \int_\mathcal{Z}k(z, .) dQ_Z(z) ||_{\mathcal{H}_k}
\end{split} 
\end{equation} 

where $\mathcal{H}_k$ is the reproducing kernel Hilbert space of real-valued functions mapping on $\mathcal{Z}$. For details on the MMD implementation, we refer to \cite{tolstikhin2017wasserstein}.  \\

\subsection{Persistence Diagram and Vietoris-Rips Complex} 
%

\begin{figure}[b] 
  \centering 
  \definecolor{dtsfsf}{rgb}{0.8274509803921568,0.1843137254901961,0.1843137254901961}
\definecolor{rvwvcq}{rgb}{0.08235294117647059,0.396078431372549,0.7529411764705882}

\begin{tikzpicture}[scale=0.9, transform shape, line cap=round,line join=round,>=triangle 45,x=1cm,y=1cm]

\draw [fill=black] (0,2.5) circle (2.5pt);

\draw [fill=black] (3,2.5) circle (2.5pt);
\draw [fill=black] (5,2.5) circle (2.5pt);
\draw [line width=0.25pt] (3,2.5) -- (5,2.5);

\fill [gray, opacity=.4] (-1,0) -- (1,0) -- (0.5,1) -- (-1,0);
\draw [fill=black] (-1,0) circle (2.5pt);
\draw [fill=black] (1,0) circle (2.5pt);
\draw [fill=black] (0.5,1) circle (2.5pt);
\draw [line width=0.25pt] (-1,0) -- (1,0);
\draw [line width=0.25pt] (1,0) -- (0.5,1);
\draw [line width=0.25pt] (-1,0) -- (0.5,1);

\fill [gray, opacity=.4] (3,0) -- (5,0) -- (4.5,1) -- (3,0);
\fill [gray, opacity=.4] (5,0) -- (4.5,1) -- (5.25,0.5) -- (5,0);
\draw [fill=black] (3,0) circle (2.5pt);
\draw [fill=black] (5,0) circle (2.5pt);
\draw [fill=black] (4.5,1) circle (2.5pt);
\draw [fill=black] (5.25,0.5) circle (2.5pt);
\draw [line width=0.25pt] (3,0) -- (5,0);
\draw [line width=0.25pt] (5,0) -- (4.5,1);
\draw [line width=0.25pt] (3,0) -- (4.5,1);
\draw [line width=0.25pt] (4.5,1) -- (5.25,0.5);
\draw [line width=0.25pt] (5,0) -- (5.25,0.5);
\draw [line width=0.25pt, dashdotted] (3,0) -- (5.25,0.5);

\end{tikzpicture} 
  \caption{A simplical complex is a collection of numerous ``simplex" or simplices, where a 0-simplex is a point, a 1-simplex is a line segment, a 2-simplex is a triangle and a 3-simplex is a tetrahedron.} \label{fig::simplex} 
\end{figure}
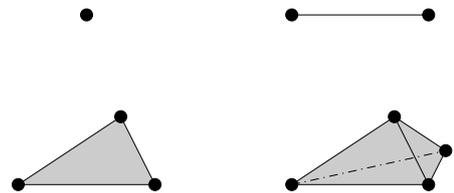

\textbf{Definition 1} Let $V=\lbrace 1, \cdots, |V|\rbrace$ be a set of vertices. A simplex $\sigma$ is a subset of vertices $\sigma \subseteq V$. A simplicial complex K on V is a collection of simplices $\lbrace \sigma \rbrace \:,\: \sigma \subseteq V$, such that $\tau \subseteq \sigma \in K \Rightarrow \tau \in K$. The dimension $n = |\sigma| - 1 $ of $\sigma$ is its number of elements minus 1. Simplicial complexes examples are represented in figure \ref{fig::simplex}.  \\

\textbf{Definition 2} Let $(X,d)$ be a metric space. The Vietoris-Rips complex $\text{VR}(X, \epsilon)$ at scale $\epsilon$ associated to $X$ is the abstract simplicial complex whose vertex set is $X$, and where $\left\lbrace x_0, x_1,...,x_k\right\rbrace$ is a $k$-simplex if and only if $d(x_i, x_j ) \leq \epsilon$ for all $0\leq i, j\leq k$.  \\ 

We obtain an increasing sequence of Vietoris-Rips complex by considering the $\text{VR}(\mathcal{C}, \epsilon)$ for an increasing sequence $(\epsilon_i)_{1 \leq i \leq N}$ of value of the scale parameter $\epsilon$ 

\begin{equation}\label{diag:seqrips} 
\mathcal{K}_1 \xhookrightarrow{\iota} \mathcal{K}_2 \xhookrightarrow{\iota} \mathcal{K}_3 \xhookrightarrow{\iota} ... \xhookrightarrow{\iota} \mathcal{K}_{N-1} \xhookrightarrow{\iota} \mathcal{K}_N. 
\end{equation} 

Applying the \textit{k-th} singular homology functor $H_k(-,F)$ with coefficient in the field $F$ \cite{hatcher2002algebraic} to \eqref{diag:seqrips}, we obtain a sequence of $F$-vector spaces, called the \textit{k-th} persistence module of $(\mathcal{K}_i)_{1 \leq i \leq N}$ 

\begin{multline} 
H_k(\mathcal{K}_1,F) \xrightarrow{t_1} H_k(\mathcal{K}_2,F) \xrightarrow{t_2} \cdots \xrightarrow{t_{N-2}} \\ \quad 
H_k(\mathcal{K}_{N-1},F) \xrightarrow{t_{N-1}} H_k(\mathcal{K}_N,F).
\end{multline} 

\textbf{Definition 3} $\forall ~i<j$, the \textit{(i,j)}-persistent $k$-homology group with coefficient in $F$ of $\mathcal{K}=(\mathcal{K}_i)_{1 \leq i\leq N}$ denoted $HP_k^{i \rightarrow j}(\mathcal{K},F)$ is defined to be the image of the homomorphism $t_{j-1} \circ \ldots \circ t_i : H_k(\mathcal{K}_i,F) \rightarrow H_k(\mathcal{K}_j,F)$.  \\ 

Using the interval decomposition theorem \cite{oudot2015Persistance}, we extract a finite family of intervals of $\mathbb{R}_+$ called persistence diagram. Each interval can be considered as a point in the set $D = \left\lbrace (x, y) \in \mathbb{R}_+^2 | x \leq y \right\rbrace$. Hence, we obtain a finite subset of the set $D$. This space of finite subsets is endowed with a matching distance called the bottleneck distance and defined as follow 

\begin{equation*} 
d_b(A,B)=\inf_{\phi : A^\prime \to B^\prime} \sup_{ x \in A^\prime}{\lVert x-\phi(x) \rVert} 
\end{equation*} 

where $A^\prime= A \cup \Delta$, $B^\prime=B \cup \Delta$, $\Delta= \lbrace (x,y) \in \mathbb{R}^2_+ \vert x=y \rbrace$ and the $\inf$ is over all the bijections from $A^\prime$ to $B^\prime$.

\subsection{Application: PHom-GeM, Persistent Homology for Generative Models} 
Bridging the gap between persistent homology and generative models, 
PHom-GeM uses a two-steps procedure. First, the minimization problem is solved for the generator $G$ and the discriminator $D$ when considering GP-WGAN and WGAN. The gradient penalty $\lambda$ in equation (\ref{eq::DWGAN}) is fixed equal to 10 for GP-WGAN and to 0 for WGAN. For auto-encoders, the minimization problem is solved for the encoder $Q$ and the decoder $G$. We use RMSProp optimizer \cite{hinton2012rmsprop} for the optimization procedure. Then, the samples of the original and generated distributions, $P_X$ and $P_G$, are mapped to persistence homology for the description of their respective manifolds. 
%
The points contained in the manifold $\mathcal{X}$ inherited from $P_X$ and the points contained in the manifold $\mathcal{\tilde{X}}$ generated with $P_G$ are randomly selected into respective batches. Two samples, $Y_1$ from $\mathcal{X}$ following $P_X$ and $Y_2$ from $\mathcal{\tilde{X}}$ following $P_G$, are selected to differentiate the topological features of the original manifold $\mathcal{X}$ and the generated manifold $\mathcal{\tilde{X}}$. The samples $Y_1$ and $Y_2$ are contained in the spaces $\mathcal{Y}_1$ and $\mathcal{Y}_2$, respectively. Then, the spaces $\mathcal{Y}_1$ and $\mathcal{Y}_2$ are transformed into metric space sets $\mathcal{\widehat{Y}}_1$ and $\mathcal{\widehat{Y}}_2$ for computational purposes. Then, we filter the metric space sets $\mathcal{\widehat{Y}}_1$ and $\mathcal{\widehat{Y}}_2$ using the Vietoris-Rips simplicial complex filtration. Given a line segment of length $\epsilon$, vertices between data points are created for data points respectively separated from a smaller distance than $\epsilon$. It leads to the construction of a collection of simplices resulting in Vietoris-Rips simplicial complex VR$(\mathcal{C}, \epsilon)$ filtration. In our case, we decide to use the Vietoris-Rips simplicial complex as it offers the best compromise between the filtration accuracy and the memory requirement \cite{chazal2017tda}. Subsequently, the persistence diagrams, $\text{dgm}_{Y_1}$ and $\text{dgm}_{Y_2}$, are constructed. We recall a persistence diagram is a stable summary representation of topological features of simplicial complex. The persistence diagrams allow the computation of the bottleneck distance $d_b(\text{dgm}_{Y_1}, \text{dgm}_{Y_2})$. Finally, the barcodes represent in a simple way the birth-death of the pairing generators of the iterated inclusions detected by the persistence diagrams.  \\

\SetAlFnt{\footnotesize} 
\SetAlCapFnt{\footnotesize} 
\SetAlCapNameFnt{\footnotesize} 
\SetKwFor{Case}{case}{}{}

\begin{algorithm}[t] 
\setstretch{1.25} 
\DontPrintSemicolon 
\KwData{training and validation sets, hyperparameter $\lambda$} 
\KwResult{persistent homology description of generative manifolds} 

\Begin{  
/*\textit{\small Step 1: Generative Models Resolution}*/ 

Select samples $\left\lbrace x_1, ..., x_n\right\rbrace$ from training set 
 
Select samples $\left\lbrace z_1, ..., z_n\right\rbrace$ from validation set 
  
With RMSProp gradient descent update ($\text{lr}=0.001, \rho=0.9, \epsilon=10^{-6}$), optimize until convergence $Q$ and $G$  \\ 

\hspace{0.25cm}\lCase{GP-WGAN and WGAN:}{using equation \ref{eq::DWGAN}} 

\hspace{0.25cm}\lCase{WAE:}{using equation \ref{eq::DWAE}} 

\hspace{0.25cm}\lCase{VAE:}{using equation described in \cite{kingma2013auto}}

\vspace{0.25cm} 

/*\textit{\small Step 2: Persistence Diagram and Bottleneck Distance on manifolds of generative models}*/ 

Random selection of samples $Y_1 \in \mathcal{Y}_1, Y_2 \in \mathcal{Y}_2$ from $P_X$ and $P_G$ 
 
Transform $\mathcal{Y}_1$ and $\mathcal{Y}_2$ spaces into a metric space set   

Filter the metric space set with a Vietoris-Rips simplicial complex $\text{VR}(\mathcal{C}, \epsilon)$ 

Compute the persistence diagrams $\text{dgm}_{Y_1}$ and $\text{dgm}_{Y_2}$ 

Evaluate the bottleneck distance $d_b(\text{dgm}_{Y_1}, \text{dgm}_{Y_2})$ 

Build the barcodes with respect to $Y_1$ and $Y_2$ 

} 

\KwRet{} 

\caption{Persistent Homology for Generative Models\label{algo::TopWGAN}} 
\end{algorithm}

\section{Experiments} \label{sec::exp} 
We empirically evaluate the proposed methodology PHom-GeM. We assess on a highly challenging data set for generative models whether PHom-GeM can simultaneously achieve (i) precise persistent homology mapping of the generated data points and (ii) accurate persistent homology distance measurement with the bottleneck distance.  \\ 

\textbf{Data Availability and Data Description} 
We train PHom-GeM on one real-world open data set: the credit card transactions data set from the Kaggle database\footnote{The data set is available at https://www.kaggle.com/mlg-ulb/creditcardfraud.} containing 284\,807 transactions including 492 frauds. This data set is particularly interesting because it reflects the scattered points distribution of the reconstructed manifold that are found during generative models' training, impacting afterward the generated adversarial samples. Furthermore, this data set is challenging because of the strong imbalance between normal and fraudulent transactions while being of high interest for the banking industry. To preserve transactions confidentiality, each transaction is composed of 28 components obtained with PCA without any description and two additional features \textit{Time} and \textit{Amount} that remained unchanged. Each transaction is labeled as fraudulent or normal in a feature called \textit{Class} which takes a value of 1 or 0, respectively.  \\ 

\textbf{Experimental Setup and Code Availability}
In our experiments, we use the Euclidean latent space $\mathcal{Z} = \mathcal{R}^2$ and the square cost function $c$ previously defined as $c(x,y)=||x-y||_2^2$ for the data points $x \in \mathcal{X}, \widetilde{x} \in \mathcal{\widetilde{X}}$. The dimensions of the true data set is $\mathcal{R}^{29}$. We kept the 28 components obtained with PCA and the amount resulting in a space of dimension 29. For the error minimization process, we used RMSProp gradient descent \cite{hinton2012rmsprop} with the parameters $\text{lr}=0.001, \rho=0.9, \epsilon=10^{-6}$ and a batch size of 64. Different values of $\lambda$ for the gradient penalty have been tested. We empirically obtained the lowest error reconstruction with $\lambda=10$ for both GP-WGAN and WAE. The coefficients of persistence homology are evaluated within the field $\mathbb{Z}/2 \mathbb{Z}$. We only consider homology groups $H_0$ and $H_1$ who represent the connected components and the loops, respectively. Higher dimensional homology groups did not noticeably improve the results quality while leading to longer computational time. The simulations were performed on a computer with 16GB of RAM, Intel i7 CPU and a Tesla K80 GPU accelerator. To ensure the reproducibility of the experiments, the code is available at the following address\footnote{The code is available at https://github.com/dagrate/phomgem}. \\


\textbf{Results and Discussions about PHom-GeM} 
We test PHom-GeM, Persistent Homology for Generative Models, on four different generative models: GP-WGAN, WGAN, WAE and VAE. 
We compare the performance of PHom-GeM on two specificities: first, qualitative visualization of the persistence diagrams and barcodes and, secondly, quantitative estimation of the persistent homology closeness using the bottleneck distance between the generated manifolds $\mathcal{\widetilde{X}}$ of the generative models and the original manifold $\mathcal{X}$.  \\ 
  
\begin{figure*}[t!] 
\centering 
\begin{subfigure}{.49\textwidth} 
  \centering 
  Persistence Diagram  \\ \vspace{0.2cm} 
  \begin{turn}{90}  
   \hspace{1.15cm} Original Sample 
  \end{turn} 
  \frame{\includegraphics[scale=0.335]{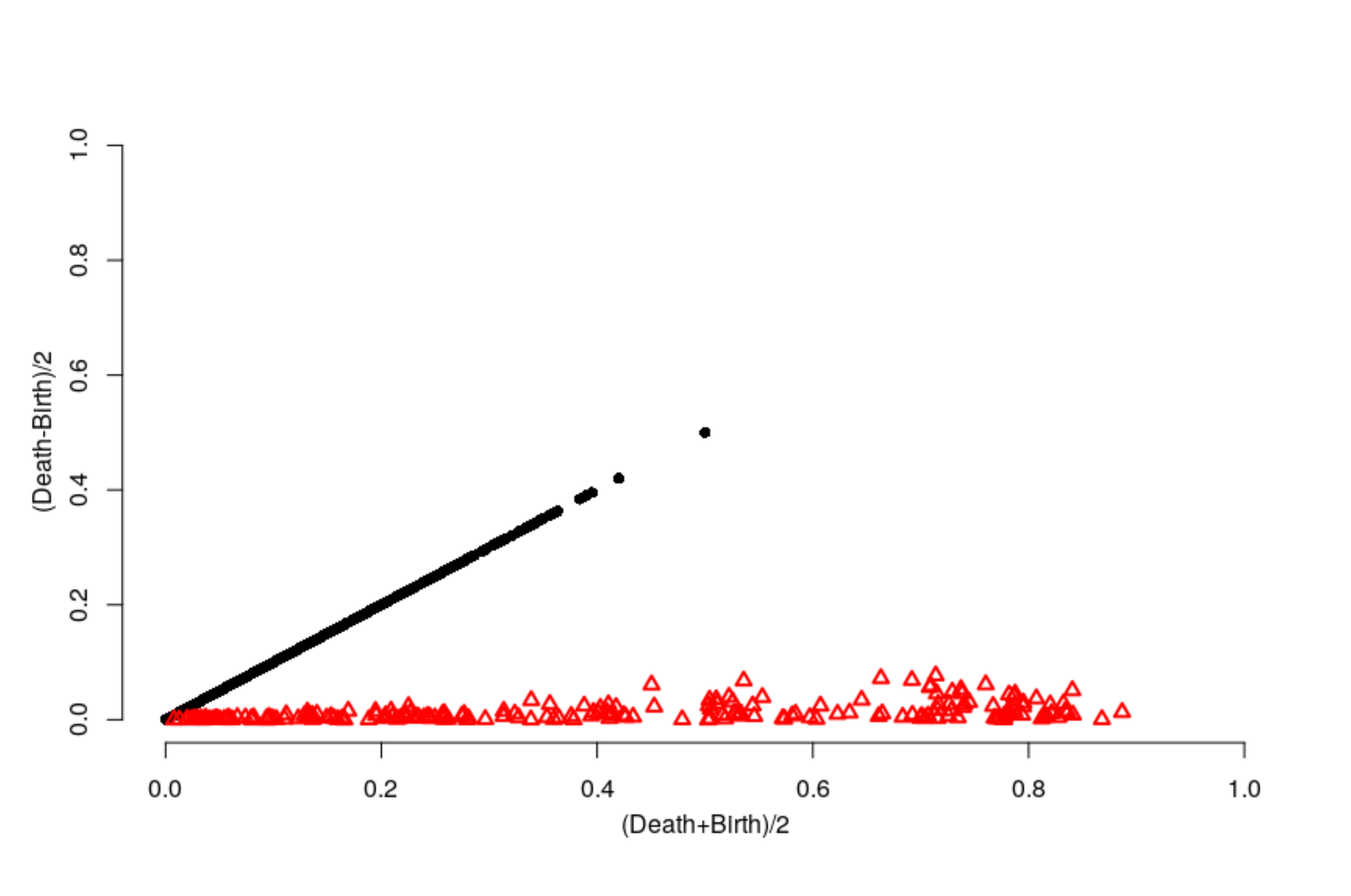}} 
\end{subfigure}\hfill 
\begin{subfigure}{.49\textwidth} 
  \centering 
  Barcodes  \\  \vspace{0.2cm} 
  \begin{turn}{90}  
   \hspace{1.15cm} Original Sample 
  \end{turn} 
  \frame{\includegraphics[scale=0.25]{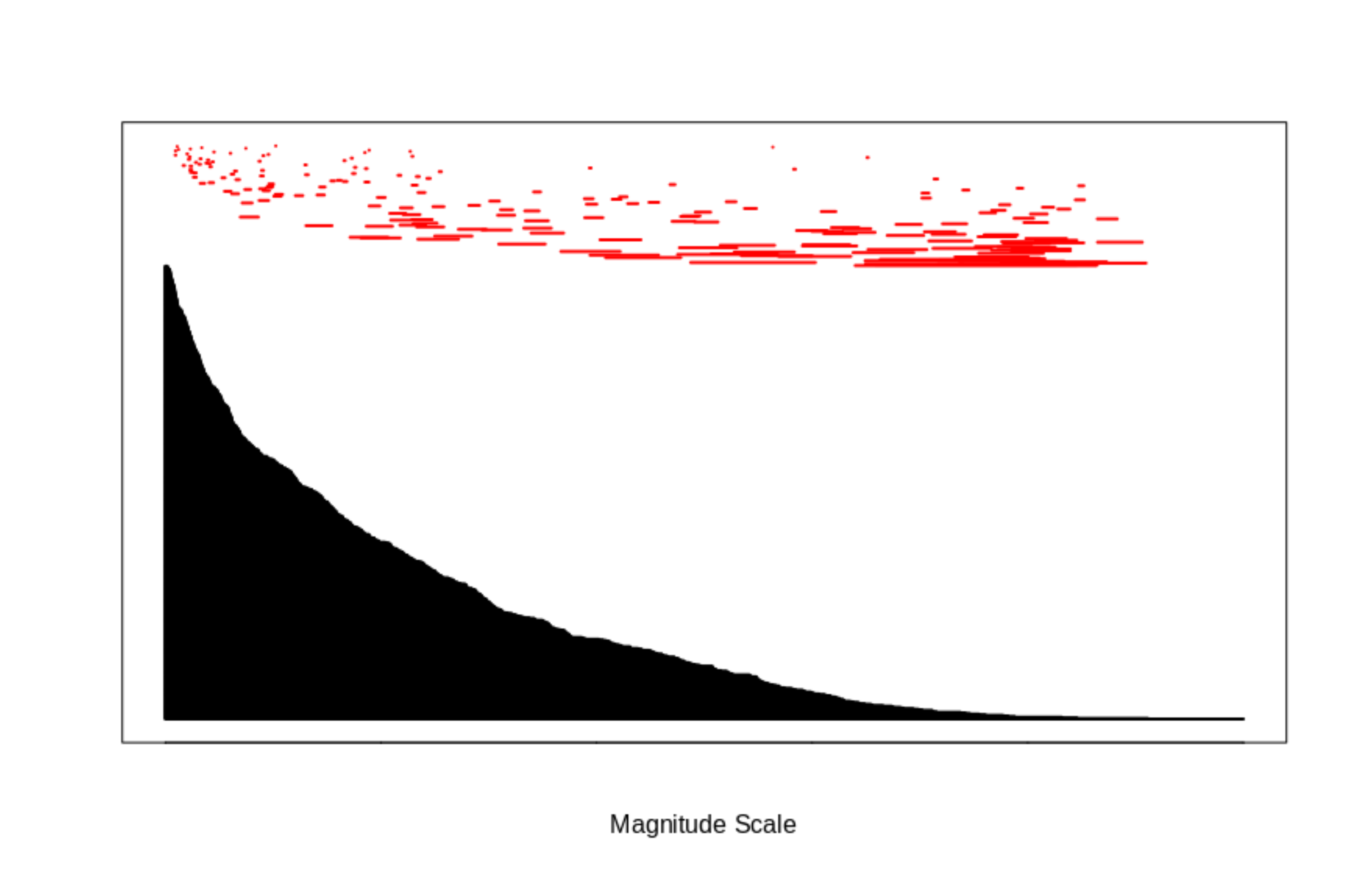}} 
\end{subfigure}\hfill 

\begin{subfigure}{.49\textwidth} 
  \centering 
  \vspace{0.2cm} 
  \begin{turn}{90}  
   \hspace{1.15cm} PHom for GP-WGAN 
  \end{turn} 
  \frame{\includegraphics[scale=0.25]{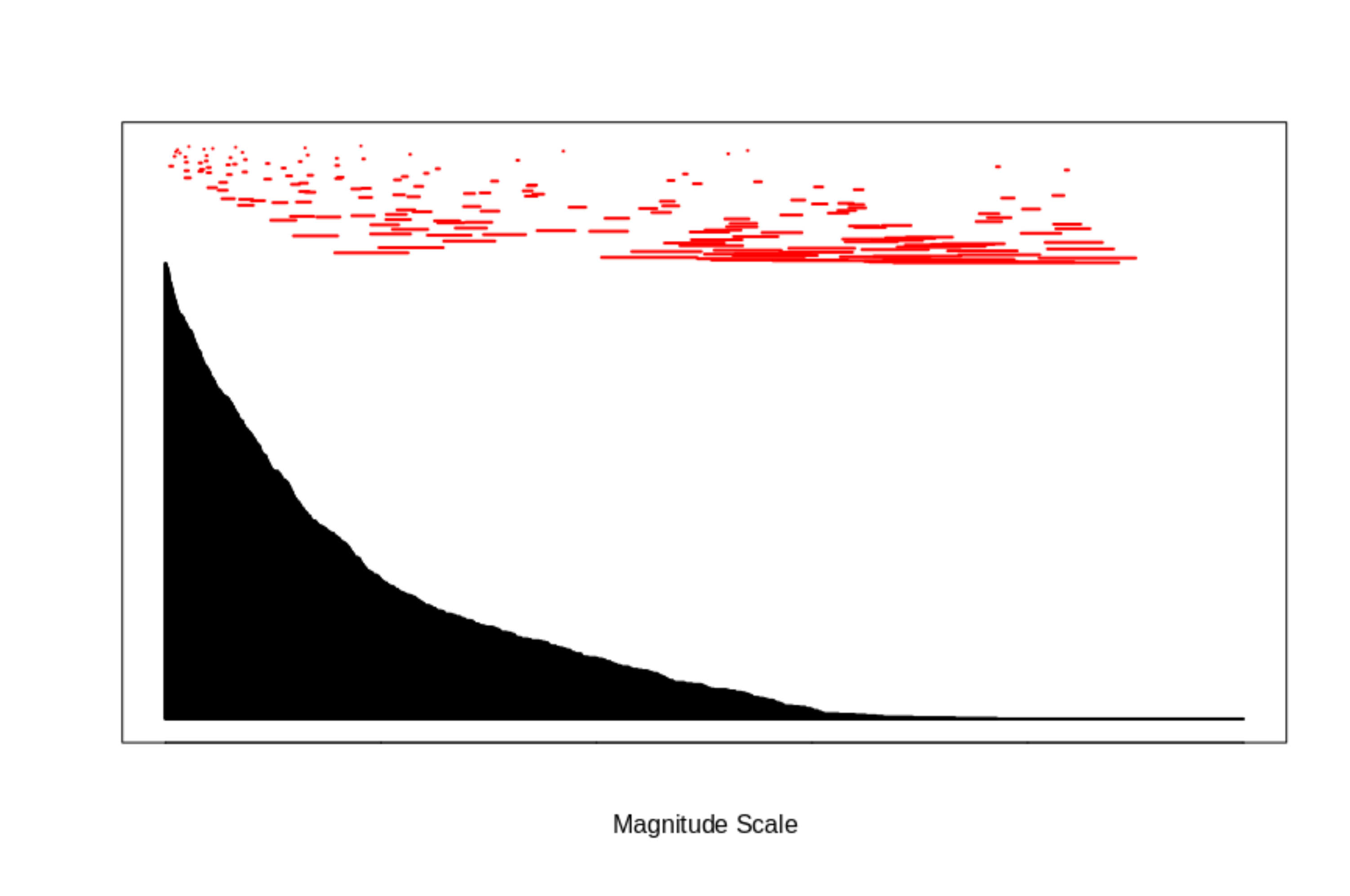}} 
\end{subfigure}\hfill 
\begin{subfigure}{.49\textwidth} 
  \centering 
  \vspace{0.2cm} 
  \begin{turn}{90}  
   \hspace{1.15cm} PHom for WGAN 
  \end{turn} 
  \frame{\includegraphics[scale=0.25]{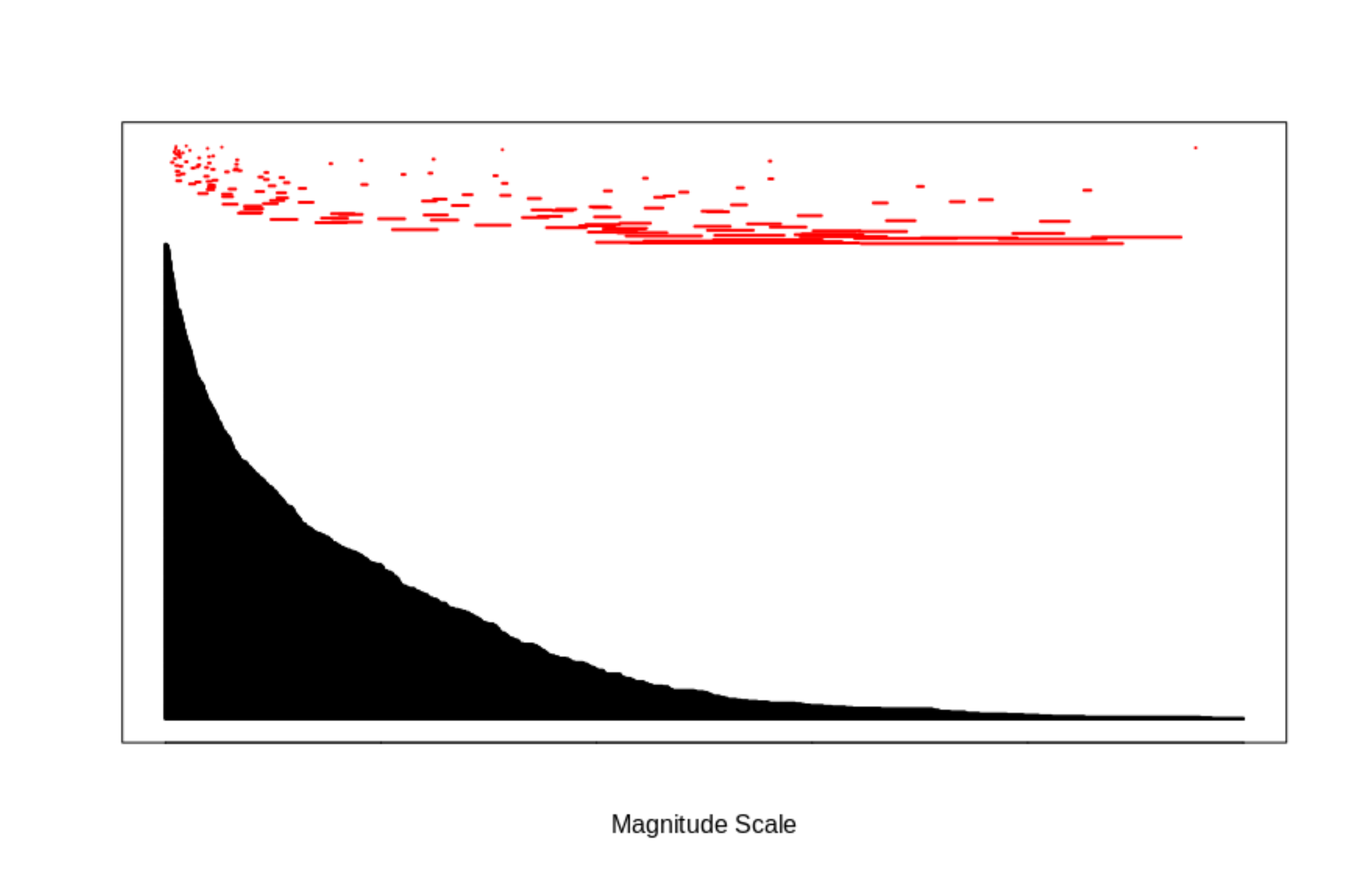}} 
\end{subfigure}\hfill 
  
\begin{subfigure}{.49\textwidth} 
  \centering 
  \vspace{0.2cm} 
  \begin{turn}{90}  
   \hspace{1.15cm} PHom for WAE 
  \end{turn} 
  \frame{\includegraphics[scale=0.25]{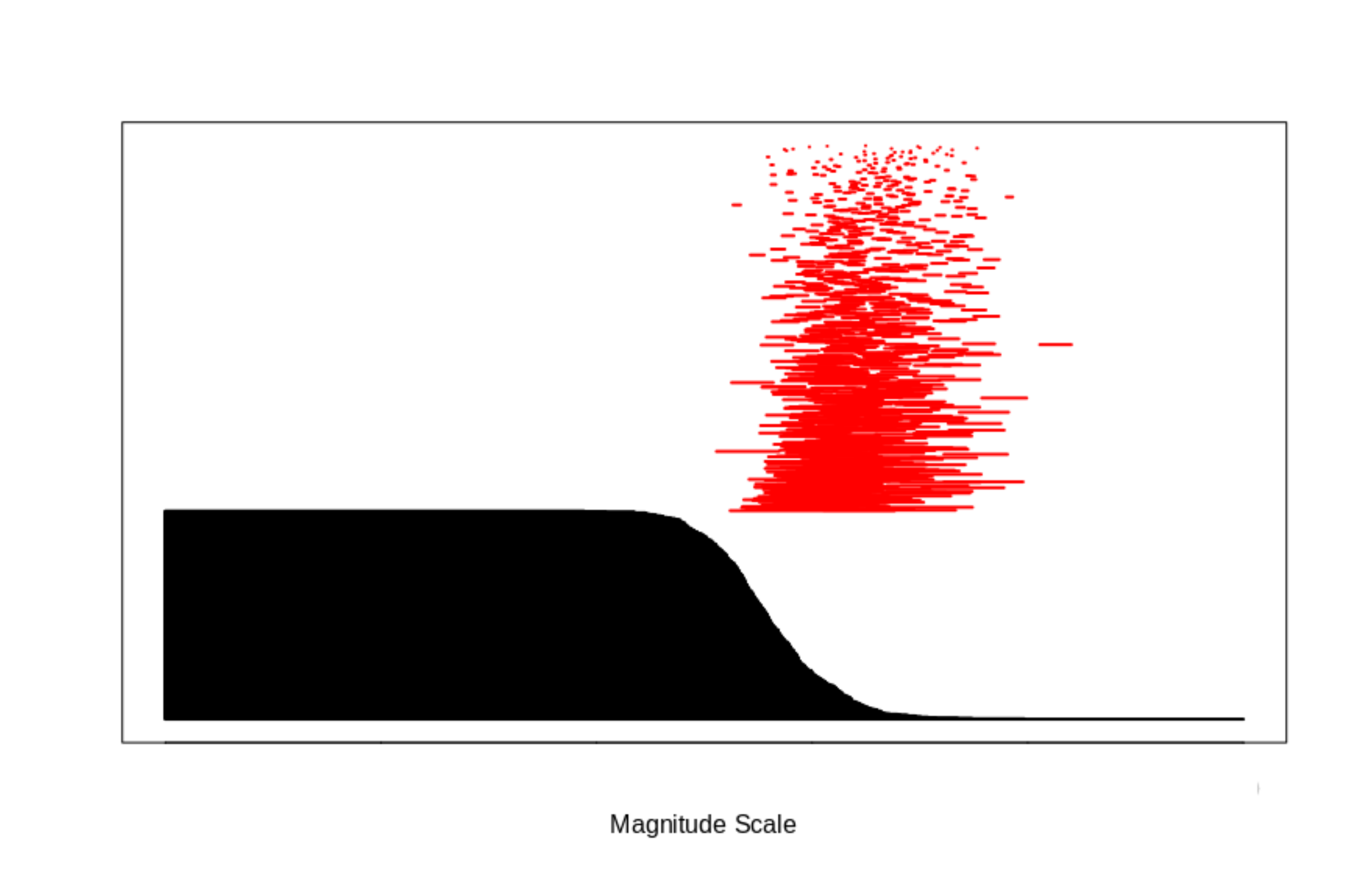}} 
\end{subfigure}\hfill 
\begin{subfigure}{.49\textwidth} 
  \centering 
  \vspace{0.2cm} 
  \begin{turn}{90}  
   \hspace{1.15cm} PHom for VAE 
  \end{turn} 
  \frame{\includegraphics[scale=0.25]{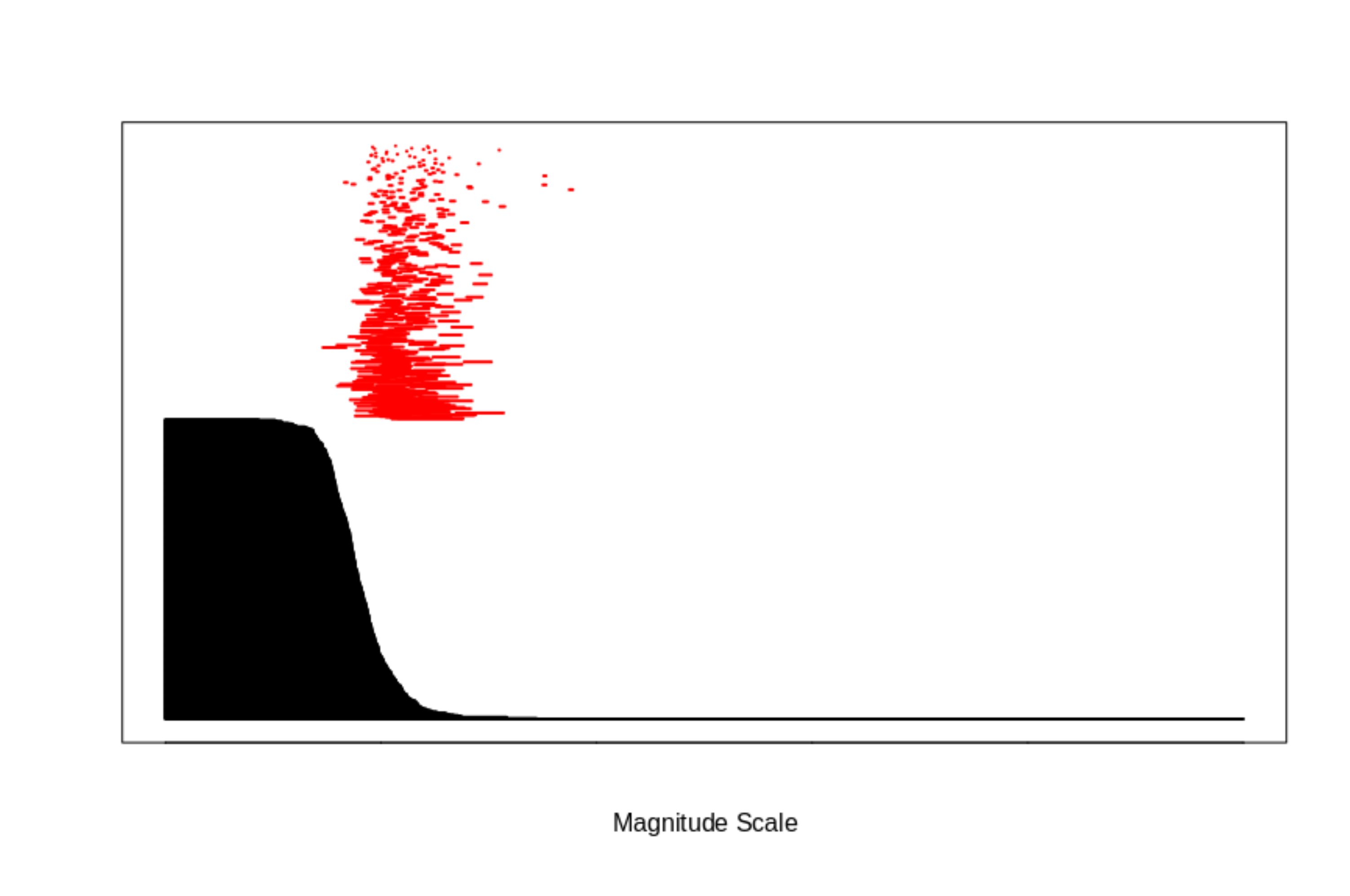}} 
\end{subfigure}\hfill 

\caption{On top, the rotated persistence and the barcode diagrams of the original sample are represented. They both illustrate the birth-death of the pairing generators of the iterated inclusions. In the persistence diagram, the black points represent the connected components of the complex and the red triangles the cycles. Below, GP-WGAN, WGAN, WAE and VAE's barcode diagrams allows to assess qualitatively, with the original sample barcode diagram, the persistent homology similarities between the generated and the original distribution, $P_G$ and $P_X$ respectively. 
} \label{fig::results} 
\end{figure*}

On the top of figure \ref{fig::results}, the rotated persistence and the barcode diagrams of the original sample $\mathcal{X}$ are highlighted. In the persistence diagram, black points represent the 0-dimensional homology groups $H_0$, the connected components of the complex. The red triangles represent the 1-dimensional homology group $H_1$, the 1-dimensional features known as cycles or loops. The barcode diagram is a simple way of representing the information contained in the persistence diagram. For the sake of simplicity, we represent only the barcode diagram of the generative models to compare qualitatively the generated distribution $P_G$ of each model with respect to the distribution $P_X$ of the original sample. The generated distribution $P_G$ of GP-WGAN is the closest to the distribution $P_X$ followed by WGAN, WAE and VAE. Effectively, the spectrum of the barcodes of GP-WGAN is very similar to the original sample's spectrum as well as denser on the right. On the opposite, the WAE and VAE's distributions $P_G$ are not able to reproduce all of the features contained in the original distribution, therefore explaining the narrower barcode spectrum.  \\

In order to quantitatively assess the quality of the generated distributions, we use the bottleneck distance between the persistent diagram of $\mathcal{X}$ and the persistent diagram of $G(Z)$ of the generated data points.
In table \ref{tab::res_ae}, we highlight the mean value of the bottleneck distance for a 95\% confidence interval. We also underline the lower and the upper bounds of the 95\% confidence interval for each generative model. Confirming the visual observations, we notice the smallest bottleneck distance, and therefore, the best result, is obtained with GP-WGAN, followed by WGAN, WAE and VAE. It means GP-WGAN is capable to generate data distribution sharing the most topological features with the original data distribution, including the nearness measurements and the overall shape. It confirms topologically on a real-world data set the claims addressed in \cite{gulrajani2017improved} of superior performance of GP-WGAN against WGAN. Furthermore, the performance of the AE cannot match the generative performance achieved by the GANs. However, the WAE, that relies on optimal transport theory, achieves better generative distribution in comparison to the popular VAE.

\begin{table}[h] 
\centering 
\caption{Bottleneck distance (smaller is better) with 95\% of confidence interval between the samples $X$ of the original manifold $\mathcal{X}$ and the generated samples $\widetilde{X}$ of the manifold $\mathcal{\widetilde{X}}$. Because of the Wasserstein distance and gradient penalty, GP-WGAN achieves better performance.} \label{tab::res_ae} 
\scalebox{0.925}{ 
\begin{tabular}{cccc} 
  \toprule 
  Gen. Model & Mean Value & Lower Bound & Upper Bound \\ 
  \midrule
  GP-WGAN & \textbf{0.0711} & \textbf{0.0683} & \textbf{0.0738}  \\ 
  WGAN & 0.0744 & 0.0716 & 0.0772  \\ 
  WAE & 0.0821 & 0.0791 & 0.0852  \\ 
  VAE & 0.0857 & 0.0833 & 0.0881  \\ 
  \bottomrule 
\end{tabular} 
} 
\end{table}

\section{Conclusion} \label{sec::ccl} 
Building upon optimal transport and unsupervised learning, we introduced PHom-GeM, Persistent Homology for Generative Models, a new characterization of the generative manifolds that uses topology and persistence homology to highlight manifold features and scattered generated distributions. We discuss the relations of GP-WGAN, WGAN, WAE and VAE in the context of unsupervised learning. Furthermore, relying on persistence homology, the bottleneck distance has been introduced to estimate quantitatively the topological features similarities between the original distribution and the generated distributions of the generative models, a specificity that current traditional distance measures fail to acknowledge. We conducted experiments showing the performance of PHom-GeM on the four generative models GP-WGAN, WGAN, WAE and VAE. We used a challenging imbalanced real-world open data set containing credit card transactions, capable of illustrating the scattered generated data distributions of the generative models and particularly suitable for the banking industry. We showed the superior topological performance of GP-WGAN in comparison to the other generative models as well as the superior performance of WAE over VAE. Future work will include further exploration of the topological features such as the influence of the simplicial complex and the possibility to integrate a topological optimization function as a regularization term.

\bibliographystyle{unsrt}
\bibliography{./zzz-mybibliography}

\begin{thebibliography}{10}

\bibitem{goodfellow2014generative}
Ian Goodfellow, Jean Pouget-Abadie, Mehdi Mirza, Bing Xu, David Warde-Farley,
  Sherjil Ozair, Aaron Courville, and Yoshua Bengio.
\newblock Generative adversarial nets.
\newblock In {\em Advances in neural information processing systems}, pages
  2672--2680, 2014.

\bibitem{goodfellow2016deep}
Ian Goodfellow, Yoshua Bengio, Aaron Courville, and Yoshua Bengio.
\newblock {\em Deep learning}, volume~1.
\newblock MIT press Cambridge, 2016.

\bibitem{kingma2013auto}
Diederik~P Kingma and Max Welling.
\newblock Auto-encoding variational bayes.
\newblock {\em arXiv preprint arXiv:1312.6114}, 2013.

\bibitem{villani2003topics}
C{\'e}dric Villani.
\newblock {\em Topics in optimal transportation}.
\newblock Number~58. American Mathematical Soc., 2003.

\bibitem{arjovsky2017wasserstein}
Martin Arjovsky, Soumith Chintala, and L{\'e}on Bottou.
\newblock Wasserstein generative adversarial networks.
\newblock In {\em International Conference on Machine Learning}, pages
  214--223, 2017.

\bibitem{gulrajani2017improved}
Ishaan Gulrajani, Faruk Ahmed, Martin Arjovsky, Vincent Dumoulin, and Aaron~C
  Courville.
\newblock Improved training of wasserstein gans.
\newblock In {\em Advances in Neural Information Processing Systems}, pages
  5767--5777, 2017.

\bibitem{tolstikhin2017wasserstein}
Ilya Tolstikhin, Olivier Bousquet, Sylvain Gelly, and Bernhard Schoelkopf.
\newblock Wasserstein auto-encoders.
\newblock {\em arXiv preprint arXiv:1711.01558}, 2017.

\bibitem{bengio2013generalized}
Yoshua Bengio, Li~Yao, Guillaume Alain, and Pascal Vincent.
\newblock Generalized denoising auto-encoders as generative models.
\newblock In {\em Advances in Neural Information Processing Systems}, pages
  899--907, 2013.

\bibitem{Edelsbrunner2002tps}
Herbert Edelsbrunner, David Letscher, and Afra Zomorodian.
\newblock {T}opological {P}ersistence and {S}implification.
\newblock {\em {D}iscrete {\&} {C}omputational {G}eometry}, 28(4):511--533,
  2002.

\bibitem{zomorodian2005computing}
Afra Zomorodian and Gunnar Carlsson.
\newblock Computing persistent homology.
\newblock {\em Discrete \& Computational Geometry}, 33(2):249--274, 2005.

\bibitem{chazal2017tda}
Fr\'ed\'eric Chazal and Bertrand Michel.
\newblock {A}n introduction to {T}opological {D}ata {A}nalysis: fundamental and
  practical aspects for data scientists.
\newblock {\em arXiv preprint arXiv:1710.04019}, 2017.

\bibitem{bousquet2017optimal}
Olivier Bousquet, Sylvain Gelly, Ilya Tolstikhin, Carl-Johann Simon-Gabriel,
  and Bernhard Schoelkopf.
\newblock From optimal transport to generative modeling: the vegan cookbook.
\newblock {\em arXiv preprint arXiv:1705.07642}, 2017.

\bibitem{hatcher2002algebraic}
A.~Hatcher.
\newblock {\em Algebraic Topology}.
\newblock Cambridge University Press, 2002.

\bibitem{oudot2015Persistance}
Steve~Y. Oudot.
\newblock {\em {Persistence Theory: From Quiver Representations to Data
  Analysis}}.
\newblock Number 209 in Mathematical Surveys and Monographs. {American
  Mathematical Society}, 2015.

\bibitem{hinton2012rmsprop}
Geoffrey Hinton, Nitish Srivastava, and Kevin Swersky.
\newblock Rmsprop: Divide the gradient by a running average of its recent
  magnitude.
\newblock {\em Neural networks for machine learning, Coursera lecture 6e},
  2012.

\end{thebibliography}

\end{document}